\documentclass[11pt]{article}
\usepackage{coling2016}
\usepackage{times}
\usepackage{url}
\usepackage{graphics}
\usepackage{graphicx}
\usepackage{multirow}

\newcommand{\tabincell}[2]{\begin{tabular}{@{}#1@{}}#2\end{tabular}}

\title{Multi-level Gated Recurrent Neural Network for Dialog Act Classification}

\author{Wei Li, Yunfang Wu \\
  MOE Key Lab of Computational Linguistics\\ School of EECS        \\
  Peking University     \\
  {\tt liweitj47@pku.edu.cn}, {\tt wuyf@pku.edu.cn} \\}

\date{}

\begin{document}
\maketitle
\begin{abstract}
In this paper we focus on the problem of dialog act (DA) labelling. This problem has recently attracted a lot of attention as it is an important sub-part of an automatic question answering system, which is currently in great demand. Traditional methods tend to see this problem as a sequence labelling task and deals with it by applying classifiers with rich features. Most of the current neural network models still omit the sequential information in the conversation.  Henceforth, we apply a novel multi-level gated recurrent neural network (GRNN) with non-textual information to predict the DA tag. Our model not only utilizes textual information, but also makes use of non-textual and contextual information. 
In comparison, our model has shown significant improvement over previous works on Switchboard Dialog Act (SWDA) task by over 6\%.
\end{abstract}


\section{Introduction}
\blfootnote{
    %
    %
    %
    
    %
    %
    \hspace{-0.65cm}  
    This work is licenced under a Creative Commons 
    Attribution 4.0 International License.
    License details:
    \url{http://creativecommons.org/licenses/by/4.0/}
}
Dialog act labelling is one of the ways to find the shallow discourse structures of natural language conversations. It represents the meaning or intention of each short sentence within a conversation by giving a tag to each sentence \cite{austin1962things,searle1969speech}. DA can be of help to many tasks, for example, the DA of the current sentence provides very important information for answer generation in an automatic question answering system. This converts a complex system into a classification problem, enabling many existing systems to fit in the problem. \newcite{allen1997draft} proposed the Dialog Act Markup in Several Layers (DAMSL) scheme to provide a top level structure for anotating dialogs, which was applied by many dialog annotation systems \cite{jurafsky1997switchboard,dhillon2004meeting}. \newcite{bunt2012iso} gave a detailed summary over the standard of dialog acts annotation in semantic annotation framework. 

Traditional methods apply classifiers with heavy human crafted features to tag the sentences. One can view each sentence in the dialog as a separate one and label it accordingly, such as the work of \cite{silva2011symbolic}, but this results in the loss of sequential information in the conversation context. \newcite{stolcke2000dialogue} used a segmented version of switchboard dialog act (SWDA) \cite{godfrey1992switchboard} with 43 tags based on the DAMSL labelling system , and proposed to use a  hidden Markov model with rich features to predict the DA of each sentence. Although their model produces relatively good results, the feature construction and tuning consume too much human effort, and also make the adaptation between tasks difficult. 

\begin{figure}
\centering
\small
\includegraphics[scale=0.35]{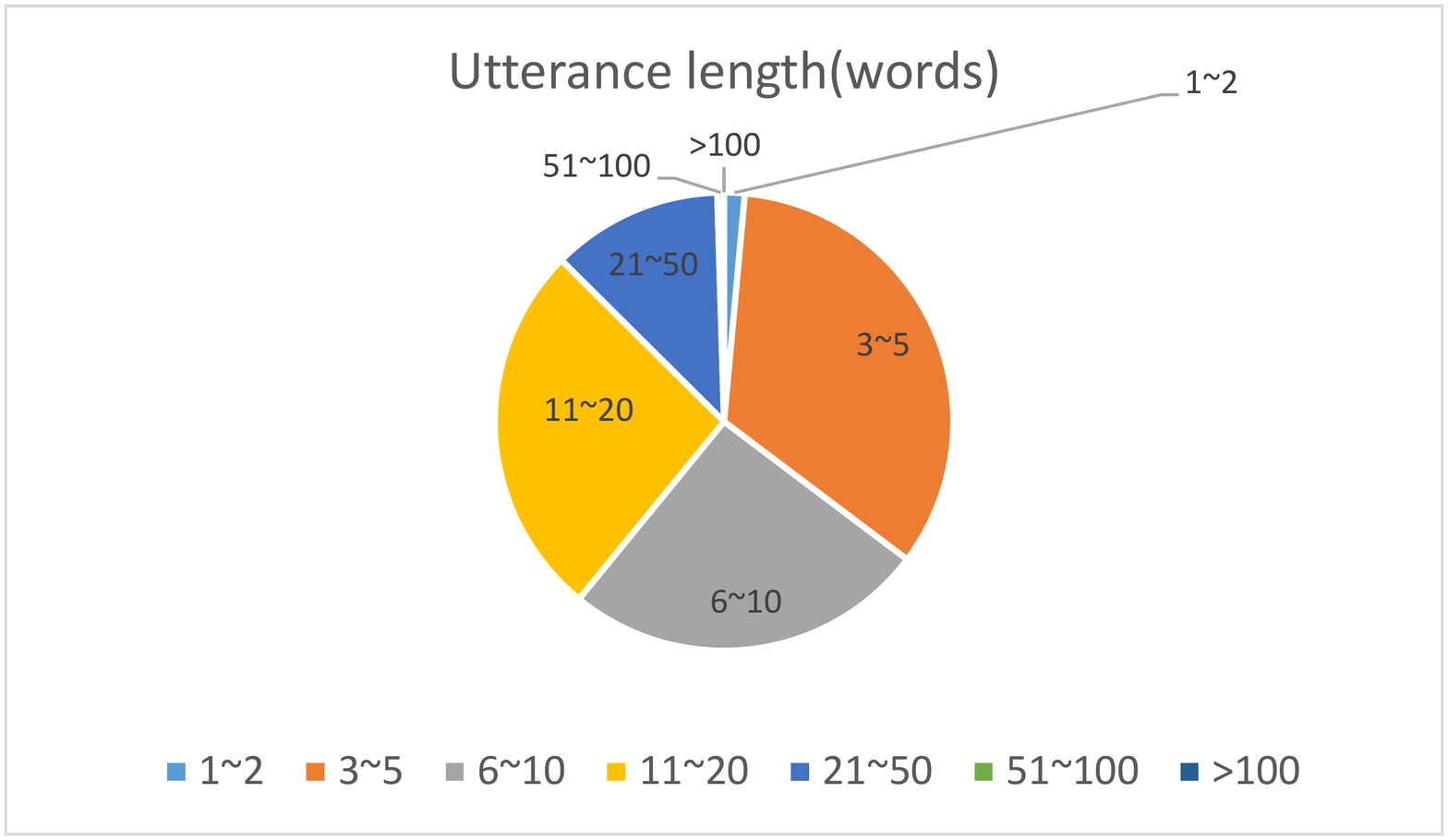}
\caption{\label{len_pic} Sentence length distribution in the SWDA corpus}
\end{figure}

Using the deep learning framework, researchers have developed various systems to deal with DA and related problems like sentiment analysis and sentence classification. One can build a simple CNN architecture like \newcite{kim2014convolutional} to do the labelling work. However, the sentences in a conversation are highly variant in length, some of which can be as short as one to two words or may even include nothing but some telephone script symbols. For example, a lot of sentences consist of nothing but "$<$\textit{laughter}$>$." and "\textit{Okay}". To be specific, in the SWDA data, 3,253 sentences consist of a single word and the length of 41.4\% sentences are under 5 words. Figure \ref{len_pic} shows the distribution of sentence lengths in detail. As is shown in the figure, most of the sentences (61\%) are under 10 words, which implies that a significant portion of the overall accuracy can be attributed to short sentences. 

Many previous models tend to do poorly on these extremely short sentences because of the lack of information. To deal with short texts, one must uncover more information, such as context sentences, to facilitate the labelling process. In fact, the most important character of DA labelling that is different from simple sentence classification is that utterances appear sequentially in a conversation. \newcite{lee2016sequential} tried to make use of historical information by feeding previous sentences in a fixed window together with the current one to a feed forward neural network. This makes a good attempt in applying contextual information. However, this approach loses long distance dependency, thus giving very little improvement when compared with the CNN baseline. \newcite{zhou2015combining} tried to capture sequential information with the Conditional random field (CRF) on the basis of a heterogeneous neural network. While their model works very well, we must also be keen to note that the RNN family models surpass CRF in sequence prediction tasks, as pointed out by \newcite{irsoy2014opinion} and \newcite{yao2014spoken}. 

Apart from textual and contextual information, non-textual information can also be considered. \newcite{hu2013multimodal} applied a restricted Boltzmann machine to combine textual and non-textual features in a community question answering problem. Their work makes good use of the non-textual features by combining them with textual features in an unsupervised manner.

To deal with the limitations of previous works, we propose a multi-level GRNN with non-textual features to predict the DAs. Our contribution can be highlighted in the following aspects:
\begin{itemize}

\item We apply a two-level GRNN to predict the DA. The low level GRNN is designed for modelling textual information of each sentence, and the top level GRNN is designed to make use of historical information in a conversation. This method produces an obvious improvement over the previous works as it automatically selects what information in the context to remember and forget.

\item We use a feed forward neural network to capture the non-textual information. Then we feed the hidden layer as sentence level non-textual information to the top level GRNN.

\item We conduct extensive experiments for DA labelling on the open SWDA corpus by exploiting different neural network models. With the new framework applied, our model achieves a significant improvement over previous works in SWDA task by over 6\% from 73.1 to 79.37.

\end{itemize}

\section{Related Work}

\subsection{Traditional methods on dialog act labelling}
Dialog act labelling was traditionally viewed as a sequence labelling or sentence modelling problem. Most of the previous works try to predict the DA by calculating the probability of each label. \newcite{reithinger1997dialogue} used a Language Model to predict the probability of a certain DA. However, as imagined, the effort to predict probability using a language model results in a severe loss of information, thereby leading to a poor result. \newcite{louwerse2006dialog} introduced n-gram features to predict the DA, which is widely used in NLP tasks. This model uncovers more information from the text, but it fails to capture long-distance dependency. \newcite{surendran2006dialog} used SVM on individual sentences then viterbi decoding to make use of contextual information in a HMM style. This model builds a rather good framework for sequential labelling, as it not only feeds each sentence to a strong classifier SVM, but also makes use of context information in a probability graph. \cite{kim2010classifying} further proposed to use CRF to deal with the problem, using both traditional bag of words features and new features such as dialog structures and dependencies between utterances. The common weakness of these methods is that they depend heavily on the features selected, and the feature construction process consumes significant human effort.

\subsection{Deep Learning models}

As deep learning becomes increasingly popular, researchers have been trying to apply deep learning frameworks to deal with natural language processing and understanding tasks, including sentence modelling, DA labelling and many other tasks. \newcite{collobert2007fast}, \newcite{collobert2008unified} and \newcite{collobert2011natural} constructed deep neural network structures for natural language processing tasks, which project one-hot word representations into distributed representations with a look-up table (or a projection layer) and build either feed forward or convolutional neural network upon them. This type of models seek to free researchers from laborious feature engineering, and allow the systems to easily adapt to different tasks. 

\newcite{kalchbrenner2014convolutional} proposed a dynamic convolution neural network with multiple layers of convolution and k-max pooling to model a sentence. As imagined, this model is computationally expensive due to the many layers. Conversely, the CNN model proposed by \newcite{kim2014convolutional} takes just one convolution and pooling layer with multi-channel word embeddings, followed by a softmax classifier. This model succeeded in many NLP tasks, such as sentence classification, sentiment analysis and so on.

Apart from CNN like architectures, researchers also applied recurrent neural network (RNN) and its variants to model sentences. Originally proposed by \newcite{elman1990finding}, RNN is expected to propagate information through time, which means one can make use of past information as latent variables. \newcite{mikolov2010recurrent} applied RNN to language modelling and got some very interesting results for word embedding. However, this vanila RNN suffers from the same problem as other deep neural networks, the problem of vanishing gradient. More specifically, gradients can either explode or vanish through time \cite{bengio1994learning}. To tackle this problem, \newcite{hochreiter1997long} proposed long short term memory (LSTM), which uses a cell with input, forget and output gates to prevent the vanishing gradient problem. This makes RNN family networks much more powerful by memorizing information from long distance. 

Recently, inspired by the gating idea, \newcite{cho2014learning} proposed another variant of RNN named gated recurrent neural network, which only uses a reset gate and a update gate, to encode and decode sentences in a translation system. As reported in \newcite{chung2014empirical}, GRNN can achieve better results than LSTM in most tasks. 

\newcite{Palangi2015Deep} proposed to sequentially take each word in a sentence, extract its information, and embed it into a semantic vector. This way, one can access the sentence level vector and use it to deal with other tasks such as information retrieval. \newcite{shen2016neural} introduced one type of attention mechanism to sentence modelling based on LSTM, they also tested their model on SWDA task, which we will reference as a baseline. Their model performed better on longer sentences by highlighting the important parts of the sentence. But, as aforementioned, the most important part of this problem is not about long sentences, but the short ones, which take the majority share of the corpus. \newcite{lee2016sequential} regarded this problem as a sequential short text classification problem, which is a good direction. However, although they tried to capture the historical information, they failed to seize long distant information in a conversation, because they only feed a fixed window to the neural network and the capability of the feed forward neural network is very limited.


\section{Our Approach}

In this paper, we propose to utilize a multi-level GRNN architecture to mine the information from both within the sentence and between the sentences. Gated recurrent neural network is a variant of the recurrent neural network (RNN). The GRNN allows information to flow over time without the problem of vanishing gradient, and is expected to memorize long distance dependency. 

Equations \ref{equ_1} to \ref{equ_2} show the method to calculate the output $h_t$ at time stamp $t$, with the input $x_t$ and history information $h_{t-1}$, which is the output at time stamp $t-1$. In each gated recurrent unit, the \textit{reset gate} (Equation \ref{equ_1}) and the \textit{update gate} (Equation \ref{equ_2}) are designed to decide which latent information is to be discarded and which is to be held. Equation \ref{equ_3} calculates the candidate unit similar to vanilla RNN unit, except that it uses a \textit{reset gate} to filter history information, and Equation \ref{equ_4} uses the \textit{update gate} and the candidate unit to get the final output unit.

In our model, we first use the low level GRNN on the scale of words to learn sentence level vector, then we use GRNN to propagate the information between sentences over time within the same conversation. To discover more information on the sentence level, we also apply a feed forward neural network to capture the non-textual information such as the length of the sentence, the index of the utterance and so on.

\begin{eqnarray}
\label{equ_1}
z_t &=& \sigma (W_z x_t + U_z h_{t-1}) \\
\label{equ_2}
r_t &=& \sigma (W_r x_t + U_r h_{t-1}) \\
\label{equ_3}
\tilde{h}_t &=& tanh (W x_t + U (r_t \odot h_{t-1})) \\
\label{equ_4}
h_t &=& (1 - z_t) h_{t-1} + z_t \tilde{h}_t
\end{eqnarray}

\subsection{\label{lrnn} Textual information}

\begin{figure}
\centering
\small
\includegraphics[scale=0.35]{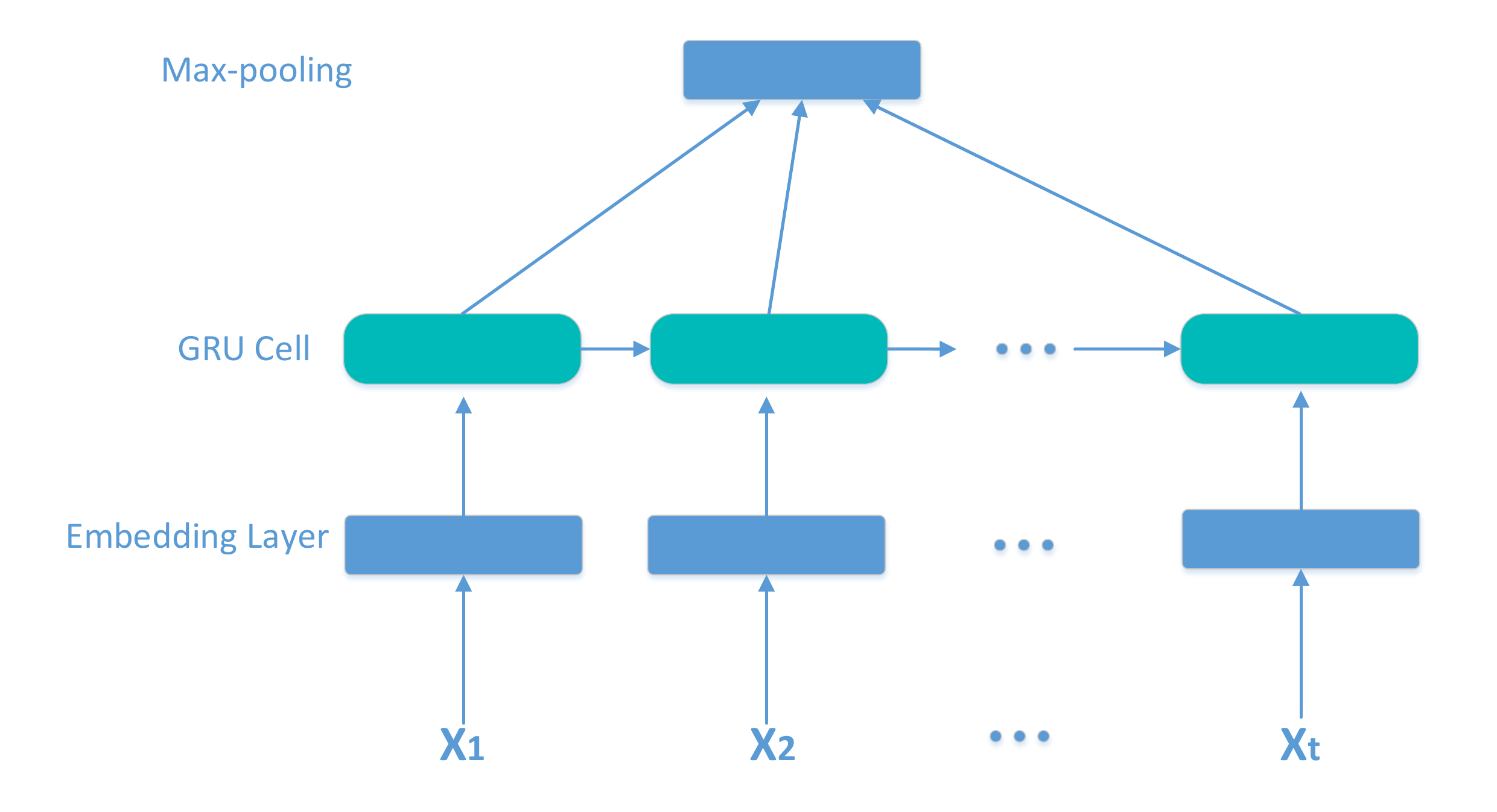}
\caption{\label{lrnn_pic} Gated recurrent neural network for sentence representation based on textual information}
\end{figure}

Textual information is the basis of our end-to-end labelling system. We use a GRNN with max-pooling to encode the sentence into a vector. 

As is shown in Figure \ref{lrnn_pic}, 
we treat each word as a separate unit. We first look up the corresponding embedding in a lookup table, which gives a matrix of $D*L$, $D$ is the dimension of word embedding and $L$ is the sentence length. Then we feed each word in the sentence into the low level GRNN, one word per time step, and then perform max pooling on the output of the GRU cells over the whole sentence.


\subsection{Non-textual information}
\label{non-textual}
Although the aforementioned low level GRNN can capture the textual information within the sentence itself, it fails to make use of information from a higher level. For instance, in our DA labelling problem, the length of sentence plays an important role in identifying the tag of sentence, because the distribution of sentence length varies between different DAs. For sentences under the label of \textit{acknowledge}, most sentences are below 10 words; whereas for sentences under the label of \textit{statement non-opinion}, the sentences have more varied length distribution. As a matter of fact, it is shown in our experiment that this sentence length feature alone gives a much better prediction than random guesses.

\begin{table}
\centering
\begin{tabular}{|c|c|c|c|c|}
\hline
caller & utterance index & sub-utterance index & act tag & text \\
\hline
A & 5 & 2 & qy & \tabincell{c}{ \{F Um, \} \{F uh, \} \\ do you
 live right in the city itself? / } \\
\hline
B & 6 & 1 & nn & No,  / \\ \hline
B & 6 & 2 & sd & I'm more out in the suburbs,  / \\ \hline
B & 6 & 3 & sd & \{C but \} I certainly work near a city. / \\ \hline
A & 7 & 1 & bk & Okay,  / \\ \hline
A & 7 & 2 & qy & \{C so \} [ ca-, + \\ \hline

\end{tabular}
\caption{\label{example} Utterance examples in SWDA corpus}
\end{table}

Feed forward neural network (FFNN) is one of the simplest form of deep neural networks, and does a good job in many tasks. In this part of the neural network, we feed four shallow non-textual features to a FFNN. We use the hidden layer as the vector representing the non-textual information of the sentence. The four features we used are listed below. To better understand the features, Table \ref{example} shows some examples from the original scripts.

\begin{itemize}
\item \textbf{Utterance index}: Conversations consist of multiple natural utterances, which are further split into lines of sentences for the convenience of tagging. Utterance index is the index of utterances, which can span multiple sentences. For example in Table \ref{example}, caller B says three sentences, and these three sentences share the same utterance index, but have different sub-utterance index. This feature may help when different acts take place in different parts of the conversation, for instance, conversations tend to begin with greetings.

\item \textbf{Sub-utterance index}: Utterances can be broken across lines, sub-utterance index gives the internal position of the current sentence in the utterance. For example, in Table \ref{example}, the 6th utterance has three sentences or sub-utterances indexing from 1 to 3. This feature helps when different acts appear in different parts of an utterance. For example, questions tend to appear at the end of each utterance.

\item \textbf{Same speaker}: This feature is a boolean feature of 0 or 1, indicating whether the identity of the speaker changes. Unlike the features above, this feature is deduced from the sub-utterance index. If the sub-utterance index is 1, then this feature is set to 1, otherwise 0.

\item \textbf{Sentence length}: As explained earlier, the length of sentence plays an important role in predicting the label. As sentence lengths vary a lot, we normalize the lengths using Equation \ref{normalization}, where $l$ is the word-wise sentence length.

\begin{equation}
\label{normalization}
l_{norm} = \frac{l- range(l)/2}{std(l)}
\end{equation}

\end{itemize}

After we have the vector for textual and non-textual information aforementioned, we concatenate them together to get a combined vector for the sentence, as shown in the lower part of Figure \ref{gru_pic}.

\subsection{Context information}
\begin{figure}
\centering
\small
\includegraphics[scale=0.35]{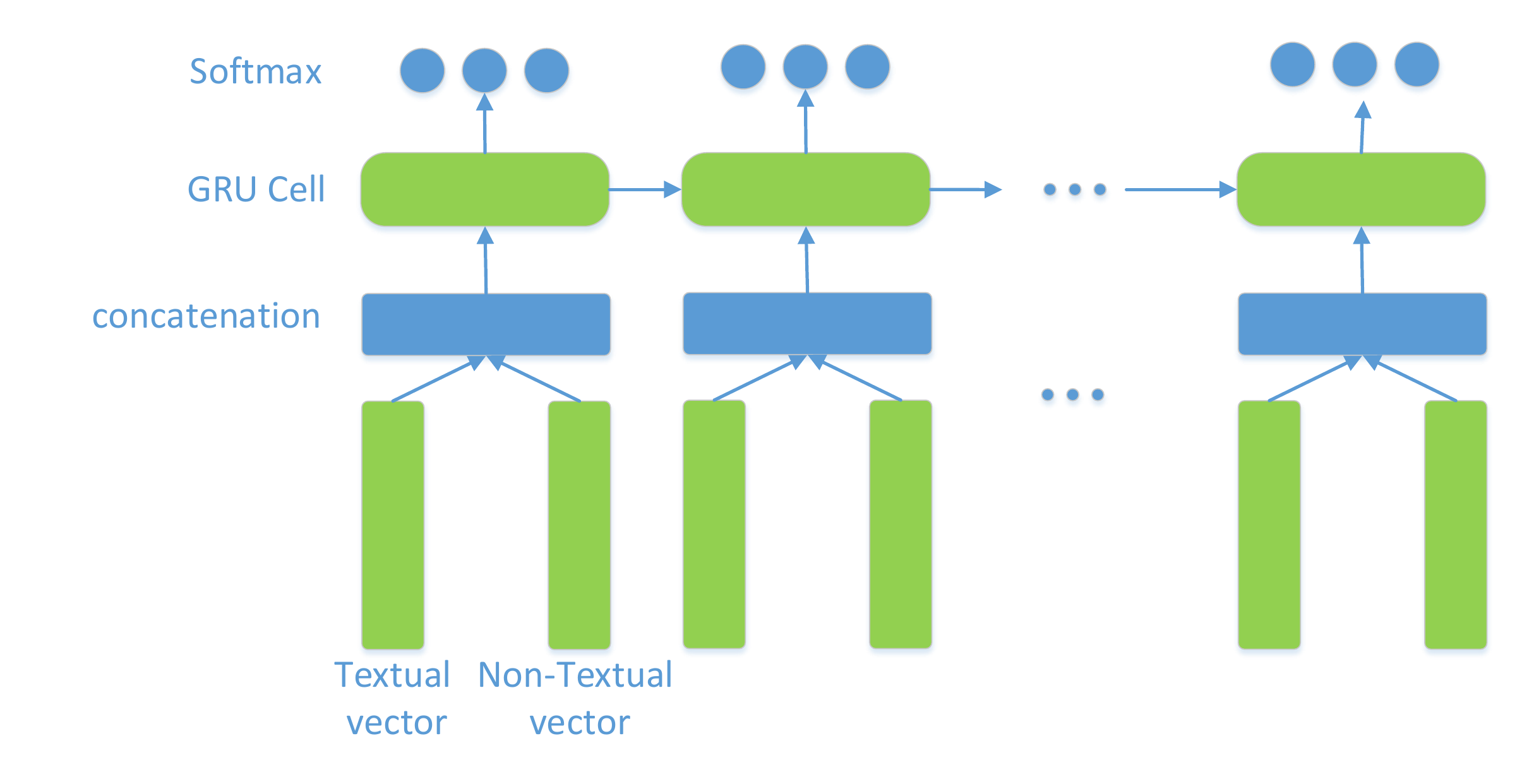}
\caption{\label{gru_pic} gated recurrent neural network on sentence feature}
\end{figure}

GRNN is designed to remember valuable information while discarding useless information. In the DA labelling problem, the segmentation of sentences is not very strict. Many sentences are very short, which makes it very difficult to classify a sentence based on only little textual information and sentence level non-textual information. Therefore, GRNN can fit this problem very well.

In our model, we try to use GRNN to capture the structure between sentences, as shown in Figure \ref{gru_pic}. This enables our model to utilize information from longer distances, unlike the structure proposed by \cite{lee2016sequential}, which uses a fixed window to capture history information. Learning distant information is crucial for the fact that the dialog turn changes with no pattern, whereas some utterances consist of a single sentence while others consist of multiple sentences, which makes it impossible to learn the words from both speakers within a fixed window, as words of one speaker in the current sentence can be distant from the last words from the other speaker.

\section{Experiment}

\subsection{Settings}
We conducted the experiment on the switchboard dialog act corpus, which extends the Switchboard-1 Telephone Speech Corpus, with turn/utterance-level dialog-act tags. The tags summarize syntactic, semantic, and pragmatic information about the associated turn. There are over 200 tags in the corpus. \cite{jurafsky1997switchboard} defines a system for collapsing them down to 44 tags. 

In our experiments, we use the same data version as \cite{stolcke2000dialogue}, where there are 1115 conversations (1.4M words, 198K utterances) in the training set, and 19 conversations  (29K words, 4K utterances) in the test set. We use the same valid set as \cite{lee2016sequential}, which consists of 19 randomly chosen conversations. \footnote{The train/validation/test splits were found at https://github.com/Franck-Dernoncourt/naacl2016}

In our experiment, we build our model upon tensorflow by \cite{tensorflow2015-whitepaper} \footnote{available in https://www.tensorflow.org}, which is a popular package developed by Google for deep learning. 

We use all the tokens of the utterances including texts and other telephone related symbols to train word embeddings with word2vec \footnote{available in https://code.google.com/archive/p/word2vec/} \cite{mikolov2013efficient}, we also set the dimension of the word embedding to 300. We use Adam stohastic optimization method \cite{kingma2014adam} to minimize the negative log-likelihood cost with fine-tuning on the word embeddings. To try to avoid the over fitting problem, we run each experiment for 10 epochs, and use the hyper parameters from the epoch with the highest validation accuracy. If not specially declared, we use rectified linear unit (relu) as the activation function.

\subsection{Baselines}

We conduct extensive experiments on the SWDA corpus by utilizing various neural network models.

\begin{itemize}
\item \textbf{CNN}: We implemented a convolutional neural network following the framework of \cite{kim2014convolutional}, we also use filters of length 2,3 and 4, and for each window length, there are 100 feature maps. So each sentence has a vector of 300 real numbers . After the convolution and max-pooling layer, there is a softmax layer to predict the DA of each sentence.

\item \textbf{non-textual}: We feed the four non-textual features to a typical three-layer feed forward neural network as described in Section \ref{non-textual}. We set the unit number of the hidden layer to 300 and use the output of the softmax layer to predict the label.

\item \textbf{CNN+non-textual}: This model is a combination of CNN and non-textual. We concatenate the pooled feature maps of CNN and the hidden layer of non-textual FFNN, and feed this new combined vector to a softmax layer to predict the label.

\item \textbf{single-level GRNN}: This model follows the description in section \ref{lrnn}. We feed the word embedding to the GRU cells, each word per cell. After we get the output of the GRU cells from each time step, we perform a max-pooling over them and get the sentence vector. Lastly, we feed the sentence vector to a softmax layer to predict the tag.

\item \textbf{single-level GRNN + non-textual}: This model combines the max-pooled sentence vector from single-level GRNN and the hidden layer of non-textual FFNN in the same way as CNN+non-textual. Then the concatenated vector is fed to a softmax layer to predict the tag.

\item \textbf{non-textual+GRNN}: We feed the hidden layer of the non-textual FFNN to a GRNN. Then we feed the output of each GRU cell to the softmax layer to predict the labels.

\item \textbf{CNN+GRNN}: We feed the sentence vector from CNN to GRNN. Then we feed the output of each GRU cell to the softmax layer to predict the labels.

\item \textbf{multi-level GRNN}: We feed the sentence vector from lower level GRNN to the upper level GRNN. Then we feed the output of each GRU cell to the softmax layer to predict the labels.

\item \textbf{CNN+non-textual+GRNN}： We feed the combination of sentence vector from CNN and hidden layer from non-textual FFNN to a GRNN. Then we feed the output of each GRU cell to the softmax layer to predict the labels.

\item \textbf{multi-level GRNN+non-textual}: This is our model in this paper. In this model, we feed the combination of sentence vector from lower level GRNN and hidden layer from non-textual FFNN to the upper level GRNN. Then we feed the output of each GRU cell to the softmax layer to predict the labels.
\end{itemize}

\subsection{Comparison with previous models}

\begin{table}
\centering
\begin{tabular}{|c|c|}
\hline
Method & Accuracy \\
\hline
\hline
Sequential short-text classification\cite{lee2016sequential} & 73.1 \\
Neural attention\cite{shen2016neural} & 72.6 \\
Our model & 79.37 \\
\hline
\end{tabular}
\caption{\label{result_compare_table} Comparison with previous state-of-the-art results}
\end{table}

Table \ref{result_compare_table} shows our result compared with other state-of-the-art results. By utilizing information from previous time stamp with GRNN, we achieved significant improvement over the previous works. As seen in Table \ref{result_compare_table}, we achieve significant improvement over both \newcite{lee2016sequential} (73.1) and \newcite{shen2016neural} (72.6) to 79.37, as we better capture both the sentence level and contextual information.

\subsection{Comparison with baseline models}

\begin{table}
\centering
\begin{tabular}{|c|c|}
\hline
Method & Accuracy \\
\hline \hline
CNN & 68.25 \\
single-level GRNN & 69.75 \\
non-textual & 43.60 \\
\hline
CNN+non-textual & 70.86 \\
single-level GRNN + non-textual & 71.90 \\
\hline
non-textual+GRNN & 48.09 \\
CNN+GRNN &  77.14 \\
multi-level GRNN & 77.65 \\
\hline
CNN+non-textual+GRNN & 78.40 \\
multi-level GRNN+non-textual & 79.37\\
\hline
\end{tabular}
\caption{\label{result_table} Results of different neural networks in our experiment}
\end{table}

Results in Table \ref{result_table} show that both CNN and single-level GRNN with textual information can give relatively good results (68.25 \& 69.75) for DA labelling problem.

Non-textual information can further improve the accuracy as they provide information about the whole sentence, instead of just individual words. This is verified by the fact that CNN+non-textual improves 2.61\% over CNN and single-level GRNN+non-textual improves 2.15\% over single-level GRNN. In fact, non-textual itself gives a surprisingly good result compared with random guess. 

It is the GRNN which captures long distance dependency from context that produces the most significant improvement to the problem. As a matter of fact, the role of GRNN is so important that GRNN based on the weak classifier non-textual FFNN improves the result by almost 5\% over the non-textual FFNN alone. GRNN on the basis of CNN improves the result by almost 10\% over the raw CNN. Altogether, our multi-level GRNN+non-textual result surpasses the CNN baseline significantly by over 11\%. 

\subsection{Analysis}
In Table \ref{result_example_table} we show the tagging results corresponding to the examples in Table \ref{example}. These results are from single-level GRNN (one of our baselines) and our final model. The sentences are selected from the first conversation in the test set.

\begin{table}
\centering
\begin{tabular}{|c|c|c|c|}
\hline
Text & standard & single-level GRNN & final model \\
\hline \hline
\{F Um, \} \{F uh, \} do you live right in the city itself? / & qy & qy & qy \\ \hline
No, / & nn & nn & nn \\ \hline
I'm more out in the suburbs, / & sd & sd & sd \\
\hline
\{C but \} I certainly work near a city. / & sd & sd & sd \\
\hline
Okay, / & bk & \textbf{fo\_o\_fw\_by\_bc} & bk \\ \hline
\{C so \} [ ca-, + & qy & \textbf{sd} & qy \\ \hline
\end{tabular}
\caption{\label{result_example_table} DA result examples of two different neural network models }
\end{table}

From the examples, we can observe that sentences with obvious characteristics can be easily recognized by both models, such as the first sentence with "\textit{do you}" is correctly tagged as "\textit{qy}" (\textit{Yes-No-Question}). However, when the sentence itself is short and ambiguous or can appear in multiple circumstances, such as "\textit{Okay}", the simpler model mistakes the "\textit{bk}" (\textit{Response Acknowledgement}) for "\textit{fo\_o\_fw\_by\_bc}" (\textit{other}), while our final model which utilizes contextual information succeeds in predicting the right tag.

\section{Conclusion}

In this paper, we describe a multi-level GRNN combined with non-textual features to deal with the dialog act labelling problem. We manage to mine multi-level information out of the conversation. Our model does a very good job on predicting short sentences. Our results surpass the state-of-the-art results significantly without much feature engineering, which makes our system easier to adapt to similar tasks. In the future, we hope to introduce attention mechanism into our model and make better use of contextual information.

\bibliographystyle{acl}
\bibliography{coling_liwei}

\begin{thebibliography}{}

\bibitem[\protect\citename{Abadi \bgroup et al.\egroup
  }2015]{tensorflow2015-whitepaper}
Mart\'{\i}n Abadi, Ashish Agarwal, Paul Barham, Eugene Brevdo, Zhifeng Chen,
  Craig Citro, Greg~S. Corrado, Andy Davis, Jeffrey Dean, Matthieu Devin,
  Sanjay Ghemawat, Ian Goodfellow, Andrew Harp, Geoffrey Irving, Michael Isard,
  Yangqing Jia, Rafal Jozefowicz, Lukasz Kaiser, Manjunath Kudlur, Josh
  Levenberg, Dan Man\'{e}, Rajat Monga, Sherry Moore, Derek Murray, Chris Olah,
  Mike Schuster, Jonathon Shlens, Benoit Steiner, Ilya Sutskever, Kunal Talwar,
  Paul Tucker, Vincent Vanhoucke, Vijay Vasudevan, Fernanda Vi\'{e}gas, Oriol
  Vinyals, Pete Warden, Martin Wattenberg, Martin Wicke, Yuan Yu, and Xiaoqiang
  Zheng.
\newblock 2015.
\newblock {TensorFlow}: Large-scale machine learning on heterogeneous systems.
\newblock Software available from tensorflow.org.

\bibitem[\protect\citename{Allen and Core}1997]{allen1997draft}
James Allen and Mark Core.
\newblock 1997.
\newblock Draft of damsl: Dialog act markup in several layers.
\newblock {\em Unpublished manuscript}, 2.

\bibitem[\protect\citename{Austin and Urmson}1962]{austin1962things}
John~Langshaw Austin and JO~Urmson.
\newblock 1962.
\newblock {\em How to Do Things with Words. The William James Lectures
  Delivered at Harvard University in 1955.[Edited by James O. Urmson.].}
\newblock Clarendon Press.

\bibitem[\protect\citename{Bengio \bgroup et al.\egroup
  }1994]{bengio1994learning}
Yoshua Bengio, Patrice Simard, and Paolo Frasconi.
\newblock 1994.
\newblock Learning long-term dependencies with gradient descent is difficult.
\newblock {\em Neural Networks, IEEE Transactions on}, 5(2):157--166.

\bibitem[\protect\citename{Bunt \bgroup et al.\egroup }2012]{bunt2012iso}
Harry Bunt, Jan Alexandersson, Jae-Woong Choe, Alex~Chengyu Fang, Koiti Hasida,
  Volha Petukhova, Andrei Popescu-Belis, and David~R Traum.
\newblock 2012.
\newblock Iso 24617-2: A semantically-based standard for dialogue annotation.
\newblock In {\em LREC}, pages 430--437. Citeseer.

\bibitem[\protect\citename{Cho \bgroup et al.\egroup }2014]{cho2014learning}
Kyunghyun Cho, Bart Van~Merri{\"e}nboer, Caglar Gulcehre, Dzmitry Bahdanau,
  Fethi Bougares, Holger Schwenk, and Yoshua Bengio.
\newblock 2014.
\newblock Learning phrase representations using rnn encoder-decoder for
  statistical machine translation.
\newblock {\em arXiv preprint arXiv:1406.1078}.

\bibitem[\protect\citename{Chung \bgroup et al.\egroup
  }2014]{chung2014empirical}
Junyoung Chung, Caglar Gulcehre, KyungHyun Cho, and Yoshua Bengio.
\newblock 2014.
\newblock Empirical evaluation of gated recurrent neural networks on sequence
  modeling.
\newblock {\em arXiv preprint arXiv:1412.3555}.

\bibitem[\protect\citename{Collobert and Weston}2007]{collobert2007fast}
Ronan Collobert and Jason Weston.
\newblock 2007.
\newblock Fast semantic extraction using a novel neural network architecture.
\newblock In {\em Annual meeting-association for computational linguistics},
  volume~45, page 560.

\bibitem[\protect\citename{Collobert and Weston}2008]{collobert2008unified}
Ronan Collobert and Jason Weston.
\newblock 2008.
\newblock A unified architecture for natural language processing: Deep neural
  networks with multitask learning.
\newblock In {\em Proceedings of the 25th international conference on Machine
  learning}, pages 160--167. ACM.

\bibitem[\protect\citename{Collobert \bgroup et al.\egroup
  }2011]{collobert2011natural}
Ronan Collobert, Jason Weston, L{\'e}on Bottou, Michael Karlen, Koray
  Kavukcuoglu, and Pavel Kuksa.
\newblock 2011.
\newblock Natural language processing (almost) from scratch.
\newblock {\em The Journal of Machine Learning Research}, 12:2493--2537.

\bibitem[\protect\citename{Dhillon \bgroup et al.\egroup
  }2004]{dhillon2004meeting}
Rajdip Dhillon, Sonali Bhagat, Hannah Carvey, and Elizabeth Shriberg.
\newblock 2004.
\newblock Meeting recorder project: Dialog act labeling guide.
\newblock Technical report, DTIC Document.

\bibitem[\protect\citename{Elman}1990]{elman1990finding}
Jeffrey~L Elman.
\newblock 1990.
\newblock Finding structure in time.
\newblock {\em Cognitive science}, 14(2):179--211.

\bibitem[\protect\citename{Godfrey \bgroup et al.\egroup
  }1992]{godfrey1992switchboard}
John~J Godfrey, Edward~C Holliman, and Jane McDaniel.
\newblock 1992.
\newblock Switchboard: Telephone speech corpus for research and development.
\newblock In {\em Acoustics, Speech, and Signal Processing, 1992. ICASSP-92.,
  1992 IEEE International Conference on}, volume~1, pages 517--520. IEEE.

\bibitem[\protect\citename{Hochreiter and Schmidhuber}1997]{hochreiter1997long}
Sepp Hochreiter and J{\"u}rgen Schmidhuber.
\newblock 1997.
\newblock Long short-term memory.
\newblock {\em Neural computation}, 9(8):1735--1780.

\bibitem[\protect\citename{Hu \bgroup et al.\egroup }2013]{hu2013multimodal}
Haifeng Hu, Bingquan Liu, Baoxun Wang, Ming Liu, and Xiaolong Wang.
\newblock 2013.
\newblock Multimodal dbn for predicting high-quality answers in cqa portals.

\bibitem[\protect\citename{Irsoy and Cardie}2014]{irsoy2014opinion}
Ozan Irsoy and Claire Cardie.
\newblock 2014.
\newblock Opinion mining with deep recurrent neural networks.
\newblock In {\em EMNLP}, pages 720--728.

\bibitem[\protect\citename{Jurafsky \bgroup et al.\egroup
  }1997]{jurafsky1997switchboard}
Dan Jurafsky, Elizabeth Shriberg, and Debra Biasca.
\newblock 1997.
\newblock Switchboard swbd-damsl shallow-discourse-function annotation coders
  manual.
\newblock {\em Institute of Cognitive Science Technical Report}, pages 97--102.

\bibitem[\protect\citename{Kalchbrenner \bgroup et al.\egroup
  }2014]{kalchbrenner2014convolutional}
Nal Kalchbrenner, Edward Grefenstette, and Phil Blunsom.
\newblock 2014.
\newblock A convolutional neural network for modelling sentences.
\newblock {\em arXiv preprint arXiv:1404.2188}.

\bibitem[\protect\citename{Kim \bgroup et al.\egroup }2010]{kim2010classifying}
Su~Nam Kim, Lawrence Cavedon, and Timothy Baldwin.
\newblock 2010.
\newblock Classifying dialogue acts in one-on-one live chats.
\newblock In {\em Proceedings of the 2010 Conference on Empirical Methods in
  Natural Language Processing}, pages 862--871. Association for Computational
  Linguistics.

\bibitem[\protect\citename{Kim}2014]{kim2014convolutional}
Yoon Kim.
\newblock 2014.
\newblock Convolutional neural networks for sentence classification.
\newblock {\em arXiv preprint arXiv:1408.5882}.

\bibitem[\protect\citename{Kingma and Ba}2014]{kingma2014adam}
Diederik Kingma and Jimmy Ba.
\newblock 2014.
\newblock Adam: A method for stochastic optimization.
\newblock {\em arXiv preprint arXiv:1412.6980}.

\bibitem[\protect\citename{Lee and Dernoncourt}2016]{lee2016sequential}
Ji~Young Lee and Franck Dernoncourt.
\newblock 2016.
\newblock Sequential short-text classification with recurrent and convolutional
  neural networks.
\newblock {\em arXiv preprint arXiv:1603.03827}.

\bibitem[\protect\citename{Louwerse and Crossley}2006]{louwerse2006dialog}
Max~M Louwerse and Scott~A Crossley.
\newblock 2006.
\newblock Dialog act classification using n-gram algorithms.
\newblock In {\em FLAIRS Conference}, pages 758--763.

\bibitem[\protect\citename{Mikolov \bgroup et al.\egroup
  }2010]{mikolov2010recurrent}
Tomas Mikolov, Martin Karafi{\'a}t, Lukas Burget, Jan Cernock{\`y}, and Sanjeev
  Khudanpur.
\newblock 2010.
\newblock Recurrent neural network based language model.
\newblock In {\em INTERSPEECH}, volume~2, page~3.

\bibitem[\protect\citename{Mikolov \bgroup et al.\egroup
  }2013]{mikolov2013efficient}
Tomas Mikolov, Kai Chen, Greg Corrado, and Jeffrey Dean.
\newblock 2013.
\newblock Efficient estimation of word representations in vector space.
\newblock {\em arXiv preprint arXiv:1301.3781}.

\bibitem[\protect\citename{Palangi \bgroup et al.\egroup
  }2015]{Palangi2015Deep}
Hamid Palangi, Li~Deng, Yelong Shen, and Jianfeng Gao.
\newblock 2015.
\newblock Deep sentence embedding using long short-term memory networks:
  Analysis and application to information retrieval.
\newblock {\em IEEE/ACM Transactions on Audio Speech {\&} Language Processing},
  24(4):694--707.

\bibitem[\protect\citename{Reithinger and Klesen}1997]{reithinger1997dialogue}
Norbert Reithinger and Martin Klesen.
\newblock 1997.
\newblock Dialogue act classification using language models.
\newblock In {\em EuroSpeech}. Citeseer.

\bibitem[\protect\citename{Searle}1969]{searle1969speech}
John~R Searle.
\newblock 1969.
\newblock {\em Speech acts: An essay in the philosophy of language}, volume
  626.
\newblock Cambridge university press.

\bibitem[\protect\citename{Shen and Lee}2016]{shen2016neural}
Sheng-syun Shen and Hung-yi Lee.
\newblock 2016.
\newblock Neural attention models for sequence classification: Analysis and
  application to key term extraction and dialogue act detection.
\newblock {\em arXiv preprint arXiv:1604.00077}.

\bibitem[\protect\citename{Silva \bgroup et al.\egroup
  }2011]{silva2011symbolic}
Joao Silva, Lu{\'\i}sa Coheur, Ana~Cristina Mendes, and Andreas Wichert.
\newblock 2011.
\newblock From symbolic to sub-symbolic information in question classification.
\newblock {\em Artificial Intelligence Review}, 35(2):137--154.

\bibitem[\protect\citename{Stolcke \bgroup et al.\egroup
  }2000]{stolcke2000dialogue}
Andreas Stolcke, Noah Coccaro, Rebecca Bates, Paul Taylor, Carol Van
  Ess-Dykema, Klaus Ries, Elizabeth Shriberg, Daniel Jurafsky, Rachel Martin,
  and Marie Meteer.
\newblock 2000.
\newblock Dialogue act modeling for automatic tagging and recognition of
  conversational speech.
\newblock {\em Computational linguistics}, 26(3):339--373.

\bibitem[\protect\citename{Surendran and Levow}2006]{surendran2006dialog}
Dinoj Surendran and Gina-Anne Levow.
\newblock 2006.
\newblock Dialog act tagging with support vector machines and hidden markov
  models.
\newblock In {\em INTERSPEECH}.

\bibitem[\protect\citename{Yao \bgroup et al.\egroup }2014]{yao2014spoken}
Kaisheng Yao, Baolin Peng, Yu~Zhang, Dong Yu, Geoffrey Zweig, and Yangyang Shi.
\newblock 2014.
\newblock Spoken language understanding using long short-term memory neural
  networks.
\newblock In {\em Spoken Language Technology Workshop (SLT), 2014 IEEE}, pages
  189--194. IEEE.

\bibitem[\protect\citename{Zhou \bgroup et al.\egroup }2015]{zhou2015combining}
Yucan Zhou, Qinghua Hu, Jie Liu, and Yuan Jia.
\newblock 2015.
\newblock Combining heterogeneous deep neural networks with conditional random
  fields for chinese dialogue act recognition.
\newblock {\em Neurocomputing}, 168:408--417.

\end{thebibliography}

\end{document}